\pdfoutput=1

\documentclass[11pt]{article}

\usepackage{ACL2023}

\usepackage{times}
\usepackage{latexsym}

\usepackage[T1]{fontenc}

\usepackage[utf8]{inputenc}

\usepackage{microtype}

\usepackage{inconsolata}

\usepackage{tikz}
\usetikzlibrary{arrows}
\usetikzlibrary{positioning}
\usepackage{tikz-dependency} 

\usepackage{amsmath}

\usepackage{cleveref}
\usepackage{nameref}
\usepackage{hyperref}
\usepackage{soul}

\usepackage{float}

\title{Conjunct Resolution in the Face of Verbal Omissions}

\author{Royi Rassin\textsuperscript{\normalfont1} \,  Yoav Goldberg\textsuperscript{\normalfont1,2} \,Reut Tsarfaty\textsuperscript{\normalfont 1}\\
\textsuperscript{1}Bar-Ilan University \, \textsuperscript{2}Allen Institute for Artificial Intelligence \\ 
  {\tt\{\href{mailto:rassinroyi@gmail.com}{rassinroyi}, \href{mailto:yoav.goldberg@gmail.com}{yoav.goldberg},
  \href{mailto:reut.tsarfaty@gmail.com}{reut.tsarfaty}\}
  @gmail.com}}

\begin{document}
\maketitle
\begin{abstract}
Verbal omissions are complex syntactic phenomena in VP coordination structures. They occur when verbs and (some of) their arguments are omitted from subsequent clauses after being explicitly stated in an initial clause. Recovering these omitted elements is necessary for accurate interpretation of the sentence, and while humans easily and intuitively fill in the missing information, state-of-the-art models continue to struggle with this task. Previous work is limited to small-scale datasets, synthetic data creation methods, and to resolution methods in the dependency-graph level. In this work we propose a {\em conjunct resolution} task that operates directly on the text and makes use of a {\em split-and-rephrase} paradigm in order to recover the missing elements in the coordination structure. To this end, we first formulate a pragmatic framework of verbal omissions which describes the different types of omissions, and develop an automatic scalable collection method. Based on this method, we curate a large dataset, containing over 10K examples of naturally-occurring verbal omissions with crowd-sourced annotations of the resolved conjuncts. We train various neural baselines for this task, and show that while our best method obtains decent performance, it leaves ample space for improvement. We propose our dataset, metrics and models as a starting point for future research on this topic.
\end{abstract}
\section{Introduction}
\begin{figure}
    \centering
    \includegraphics[width=0.8\columnwidth]{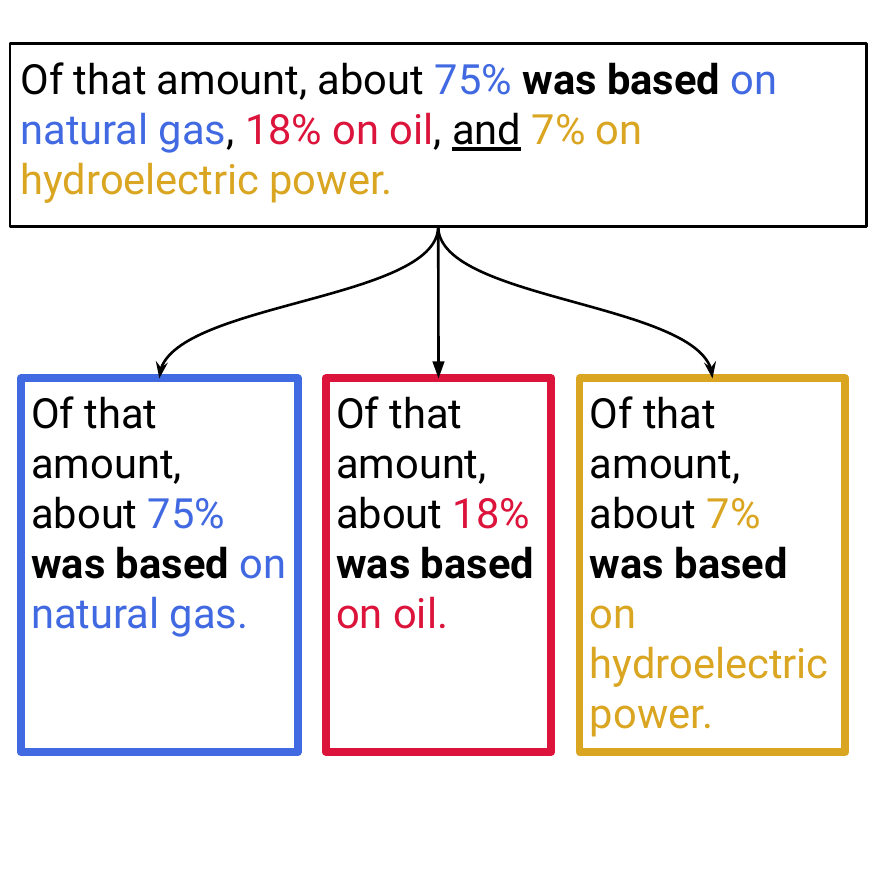} 
    \caption{A demonstration of the conjunct resolution task. At the top, the input sentence with the omitted words in bold and the coordinating element is underlined. Below, the expected rewrite, consisting of a single fully-specified standalone statement for each of the conjuncts. The rewrite must not include the coordinating element.}
\label{fig:rewrites}
\end{figure}
Natural language is economic, and many elements in the message are not spelled out but omitted by speakers,
 left for the receiver of the message to complete. The hearer, either a human or an algorithm, should then
complete the missing information and recover the intended meaning. This kind of omission and recovery occurs at all levels of conversation,
from syntax to pragmatics, and is performed naturally and intuitively by humans, to the extent that they often don't even realize that something was missing in the message they received.

\begin{figure}[h!]
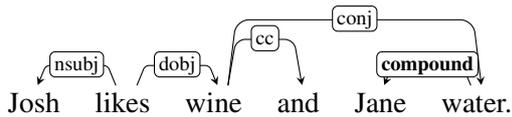

    \centering
    \tikzset{every node/.style={align=center}}
    \begin{dependency}
    \begin{deptext}[column sep=0.5em]
      Josh \& likes \& wine \& and \& Jane \& water. \\
    \end{deptext}
    \depedge[edge height=1.5ex]{2}{1}{nsubj}
    \depedge[edge height=1.5ex]{2}{3}{dobj}
    \depedge[edge height=3.5ex]{3}{4}{cc}
    \depedge[edge height=5ex]{3}{6}{conj}
    \depedge[edge height=1.5ex]{6}{5}{\textbf{compound}}
  \end{dependency}
  \caption{Compound relation indicating a misinterpretation of the sentence "Josh likes wine and Jane water" instead of "Josh likes wine and Jane likes water"}
  \label{graph:jane_water}
\end{figure}

An important class of omission phenomena --- and the focus of this work --- involves verbs and their arguments, especially around coordination structures. A verb  and some of its arguments that appear in an initial clause may be omitted in subsequent clauses.
For example, in the sentence ``Josh likes wine and Jane water'', the verb \emph{likes} is omitted from the phrase ``Jane likes water'' and has to be inferred. We find that state-of-the-art syntactic parsers \citep{spacy2, trankit} fail consistently even on toy examples such as this one, and assign a structure in which ``Jane water'' is interpreted as a noun-compound which is the object of Josh's liking, together with water (Figure~\ref{graph:jane_water}). 
Downstream applications, such as, Google translate\footnote{\url{https://translate.google.com/}, accessed on Jan 18 2023.} also fail. Google translate  translates the above sentence to French as ``Josh aime le vin et l'eau de Jane''. When back-translated to English using the same engine, it results in ``Josh likes Jane's wine and water''.
Significantly more complex sentences involving verbal omissions are of course also possible, and  NLP systems routinely fail around them.

Linguistically, such verbal omissions are studied under the terms \emph{ellipsis} and \emph{gapping}, and the research broadly aims at categorizing the verbal omission cases into sub-categories, carefully documenting the condition under which different omissions can or cannot occur \citep{hudson1989gapping, Ross1967aGapping, jackendoff1971gapping}. For language technology purposes, however, we would like to focus on detection, resolution and usability, not categorization. The need to recover empty elements automatically and robustly in large-scale calls  for a unifying approach, by which to consider the different cases of verbal omissions as instances of a single phenomenon, that can be effectively addressed by a broad-coverage NLP system.

 Due to failures of even language technology applications on such constructions, in this paper, we suggest that these verbal omission phenomena should be addressed explicitly by systems, rather than hoping it will be caught on as a side product of end-to-end neural training. 
To this end, our aim  is to establish a verbal-omission recovery task, and to create a supporting large-scale corpus documenting such verbal omissions in naturally occurring English sentences, together with automatic annotation models that recover  the implicit information and make it explicit.  

How should a system that resolves such missing information look like? What are its outputs? In the syntactic parsing and treebanking literature, there has been an ongoing debate regarding the best way to represent recovery of omissions in coordinated structures \citep{treebank_guidelines, ud2, gapping_ud2,  conjunction_reduction, vpe_automatic_parsed, sluicing_1, sluicing_2, pseudogapping, ficler_goldberg_acc, gapping_short_paper, parseMeIfYouCan, mindTheGap}. 

However, none of the solutions are fully satisfactory. First, all of them are highly technical in nature, and require  significant linguistic expertise in order to even understand the notation. Often they assume some formal tree or graph notation, which makes their adoption unlikely by people who work on NLP systems but who may lack the specific syntactic expertise. Moreover, how can one feed such theory-loaded graph-based annotation into NLP systems for downstream, user-facing tasks? 

To mitigate this, our proposed solution is strictly within the text-to-text paradigm, and involves a re-writing task where the input is a sentence exhibiting a coordinated structure, and the output is the set of sentences which contain the same information, but where the implicit information is made explicit. We illustrate an input and output example in Figure~\labelcref{fig:rewrites}. As we further discuss below, this text-enrichment approach has several appealing properties: first, it is natural and easily comprehensible to any competent speaker of the language. Second, this  caters for both large-scale annotation efforts and task adoption by potential users. Lastly, it produces an output which can trivially be fed into any language processing system which takes natural language text as input.

We collect a corpus of 10,206 sentences involving a wide range of verbal omissions  in coordinated structures, annotated via this text-rewriting task to recover the missing elements. Using this corpus, we conduct  experiments on the omission-recovery task. Training a T5-based model to perform the task shows that the performance saturates after seeing only 10\% of the data, but the accuracy of the  neural model is far from being perfect, leaving ample room for future modeling and improvements.\footnote{We make our code and dataset publicly available at \url{https://github.com/RoyiRa/conjunct-resolution-task}}
\section{Related Work}
Previous research on verbal omissions in coordination structures  has classified such phenomena  into (at least)  six categories: (1) conjunction reduction \citep{conjunction_reduction}; (2) gapping \citep{Ross1967aGapping, jackendoff1971gapping, hudson1989gapping}; (3) VP Ellipsis \citep{vpe_automatic_parsed}; (4) sluicing \citep{sluicing_1, sluicing_2}; (5) pseudo-gapping \citep{pseudogapping}; and (6) argument clusters. All of these phenomena are forms of ellipsis and are considered as ways to use language efficiently. 

How do NLP systems cope with such complex syntactic phenomena? Work around gapping in the syntactic parsing community has focused on representation schemes for this phenomena in dependency graphs. Various representation has been proposed. For instance, the Universal Dependencies (UD) framework \citep{ud2} introduces the concept of an "orphan" dependency to indicate the presence of an ellipsis, and \citep{gapping_ud2} provides a detailed analysis of how gapping constructions could be represented using UD. Related work \citep{sebastianGapping, mindTheGap, parseMeIfYouCan} utilize such schemes by promoting a new head of the clause in cases of gapping and attaching all  remnants to it.

However, having a representation schema for a construction does not mean that an automatic parser is able to accurately predict it.
To bridge this gap (no pun intended), 
several efforts have been made to increase the training data size for these constructions
by enriching existing data or creating artificial datasets. \citet{parseMeIfYouCan} proposed data enrichment methods that utilize existing annotated parse trees to mimic the structure of elliptical constructions, and \citet{mindTheGap} trained a parser on artificial elliptical treebanks,  
achieving an F1 score of 36\% on a small dataset (an improvement relative to prior work). Moreover, \citet{sebastianGapping, gapping_short_paper} proposed a reconstruction algorithm at the dependency graph level, but relied on an oracle to identify the gapped sentences. 

Despite these efforts, current approaches do not fully address the issue of verbal omissions in coordination structures, and remain scattered. Additionally, these methods are not suitable for large-scale applications and are not ready for use in downstream tasks.
In this work we take a more realistic, consistent and unified approach, wherein all verbal omissions are treated within the same framework, for which we provide a gold benchmark, a crowdsourcing interface, models for the tasks demonstrating feasibility and efficacy. 
\section{The Conjunct Resolution Task} 
\label{resolution_task}
\subsection{Desiderata} \label{sec:desiderata}
We seek a unifying approach to handle many types of verb-related omissions in coordination structures, and which is targeted primarily at users of language technologies. Concretely, our approach should allow for (1) an annotation scheme that is scalable to multiple annotators while maintaining quality; (2) a comprehensible task to non-linguists that is simple enough that no specialized linguistic expertise is required by a user of the resulting annotations; (3) amenable to automatic annotation by models; (4) useful in the context of a downstream language processing applications. Additionally, it would be preferable if the approach is language-agnostic, and that the approach is not constrained by any particular linguistic theory.

\subsection{The Task}
In order to meet the aforementioned desiderata 
(Section~\ref{sec:desiderata}), our approach steps away from traditional syntactic representation and instead represents the output as natural language text, which is an enriched version of the input text. This offers several advantages. First, the resulting annotation task is intuitive for annotators as they now need to read a sentence and rewrite it, which is a natural and familiar setting; second, language models are designed to learn from, operate on, and produce natural text representation; finally, the output can be consumed by any process that takes text as input, so applications can benefit from the enrichment without requiring a change in their design.

Intuitively, such a task could be to take a sentence with missing information, and rewrite it by completing in all the missing verbs and their arguments, e.g. rewriting ``Josh likes wine and Jane water'' to ``Josh likes wine and Jane likes water''. However, we found this task to be notoriously hard to explain both to non-linguists (who did not realize a verb was missing in this sentence in the first place) and to linguists (who also do not identify some of the verbs as ``missing'', for example in ``Jane likes wine and water'', which is not technically a gapping construction but a verb taking a coordinated NP as its object --- but which we aim to reconstruct as two distinct conjuncts\footnote{In this work we use the word ``conjuncts'' to mean expressions linked by the conjunctions ``and'', ``or'', or ``but''.}  nonetheless).

Instead,
we build on the {\em split-and-rephrase} paradigm \citep{splitAndRephrase}, which involves breaking down a complex sentence into smaller, simpler sentences. Specifically, we propose to decompose sentences that potentially involve verbal omissions around coordination into a set of independent sentences, that together capture the meaning of the original sentence, and do not add to it. In contrast to the original split-and-rephrase work, where no information is actually missing, 

the rewritten sentences make  the implicit arguments explicit in each conjunct, and when they are taken together, these sentences retain the meaning of the original complex sentence. 

Concretely, we define the conjunct resolution task as follows: given a sentence containing one of the conjunctions ``or", ``and", or ``but" as input, the sentence has to be rewritten into a set of sentences while adhering to the following constraints:\begin{itemize}
    \item[] 
(1) the set of sentences must not include the marked conjunction;\\ (2) the sentences should introduce a minimal number of new content words;\\(3) the sentences set  should preserve the meaning of the original sentence and not add to it; 
\end{itemize}
If it is not possible to rewrite the sentence under the preceding constraints without altering its meaning, the sentence should be left unchanged.

The first constraint drives the annotation: by not allowing to use the conjunction, the sentence must be split, and all the verbs and their arguments must be spelled out. The two other constraints keep the resulting sentence set both minimal and complete.

To illustrate our task, consider the following sentence. The underlines indicate omitted elements, and are not part of the input. The focused conjunction "and" is marked in bold:
\begin{itemize}
\item
``\emph{As of January 2013, The Times has a circulation of 399,339, The Sunday Times \_\_ of 885,612, \textbf{and} The New York Times \_\_ of 9,512,132.}''
\end{itemize}
\noindent The sentence is rewritten into a set of three sentences as each clause describes a unique event:

\begin{enumerate}
\renewcommand*\labelenumi{(\theenumi)}
 \item As of January 2013, The Times has a circulation of 399,339.
 \item As of January 2013, The Sunday Times has a circulation of 885,612.
 \item As of January 2013, The New York Times has a circulation of 9,512,132.
\end{enumerate}

A core challenge of this task is to rewrite the sentence to a correct number of sentences while faithfully retaining the meaning of each clause (for instance, including the opening span "As of January 2013," is crucial in retaining the overall meaning of the sentence, however, it should not be a  sentence on its own, as it is not a coordinated clause). 

The closing clause refers to sentences that can not be rewritten to independent sentences, but still contain verbal omissions. For instance, consider \begin{itemize}
    \item 
``\emph{Amla made 133 \textbf{and} Roussow \_\_ 132 with the pair combining to put on 247 for the third wicket.}''
\end{itemize}
while this is indeed a case of verbal omissions, the span ``\emph{with the pair combining to put on 247}'' binds the two clauses together in a way that will lose its meaning if rewritten to a set of two sentences. 

This paradigm fits our objective well and addresses the desiderata we put forth in Section~\ref{sec:desiderata}, as rewriting a sentence to a set of sentences indirectly resolves the verbal omissions. It is amicable to non-linguist annotators and straightforward to scale while maintaining quality, as there is little room for variance when breaking down a complex sentence. Finally, users of the process can clearly and intuitively understand its intended behavior and can analyze its correctness, without requiring any linguistic training.

\section{Data Collection Process} \label{sec:collection_process}
We collect a dataset of 10,206 examples of a wide array of omission cases, which can serve for both training and evaluation. We aim for the collected sentences to cover a wide range of omission cases.
The underlying data was sourced from three publicly available datasets: SQuAD 2.0, Dailymail, and CNN \citep{squad2dataset, cnn_Dailymail}, with a roughly equal proportion of sentences from each one. 
Our proposed collection protocol involves two steps: (1) automatic collection of  sentences that are likely to contain interesting omission phenomena; (2) manual annotation of the sentences via crowd-sourcing. 

\subsection{Sentences Collection}\label{sentence_collection}
Instead of trying to explicitly target and identify specific types of verbal omissions, we instead rely on the observation that these verbal omission constructions affect both manual and automatic syntactic analysis in various ways, either due to constructions that are hard or impossible to represent, or due to parsing mistakes. We thus do not look {\em directly} for specific verbal omissions, but rather, identify their side-effects as manifested in the graph outputs of a dependency parser. For example, here:

    \tikzset{every node/.style={align=center}}
    \begin{dependency}
    \begin{deptext}[column sep=0.5em]
      32\% \& had \& brown \& and \& 21\% \& black. \\
    \end{deptext}
    \depedge[edge height=1.5ex]{2}{1}{nsubj}
    \depedge[edge height=1.5ex]{2}{3}{dobj}
    \depedge[edge height=3.5ex]{2}{4}{cc}
    \depedge[edge height=5.0ex]{2}{6}{conj}
    \depedge[edge height=1.5ex]{6}{5}{\textbf{nsubj}}
  \end{dependency}
the omission of \emph{had} manifests in two ``suspicious'' structures: a \textit{conj} relation between two different part-of-speech tags, and an \textit{nsubj} dependent of a word which is not a verb.

We collect cases based on 21 such patterns, applied to sentences that include coordination.

By considering a sample of the sentences identified in this manner, we verified that roughly 92\% of the resulting cases are indeed non-trivial verbal omission cases.

\subsection{Annotation and Curation}
We devise a scalable annotation procedure that can be performed by non-expert annotators.\footnote{We rely on a pool of trained crowd workers in the controlled crowd-sourcing paradigm \cite{roit_controlled}.}
The procedure is based on the conjunction resolution task (Section~\ref{resolution_task}), which does not require annotators to possess advanced linguistic knowledge. Instead, it relies on their intuitive understanding of language. 

\paragraph{Crowdsourcing Infrastructure.} We set up an Amazon Mechanical Turk (AMT) task in which workers were given a coordination structure with suspected omissions and a highlighted conjunction, and were asked to rewrite the sentence into multiple independent sentences, according to our task's rules. Each AMT assignment (known as a "HIT") begins with a brief description of the task and two examples: one example of a rewritable sentence and the other of a not rewritable sentence. Workers were also given an option to view five rewritable and five non-rewritable examples with detailed explanations. 

A HIT consisted of seven pairs consisting of a coordination sentence and a highlighted conjunction in it. The annotators were requested to rewrite the sentences in the order in which the clauses are read in the sentence. We encouraged annotators to review their work by joining the set of sentences with the highlighted conjunction and comparing the meaning between the input and the sum of their sentences. Moreover, when annotators submitted a sentence unchanged, they were prompted to explain why. This not only facilitated critical thinking on their part, but also allowed for potential revisions to their annotation and provided valuable insight on the data, being a useful resource on its own.

To ensure the quality of the annotations, we implemented several checks: (1) We ensured that no two sentences in the rewritten set were identical. (2) We verified that the highlighted conjunction was not present in any of the sentences in the set. (3) We confirmed that no new content words were added, while still allowing for inflectional variations in verb and noun forms to maintain grammatical accuracy. Additionally, to handle unexpected cases and gain further understanding of the task, annotators were given the option to indicate uncertainty in their annotation and to specify if the sentence was a ``long list'' requiring more than ten rewrites. In case of the latter, sentences were removed. Annotators were also given the option to provide any feedback. \cref{app:crowdsourcing_task} shows the user interface of this task.

\paragraph{Annotations Consolidation.} The final annotations were determined by majority agreement among annotators on the number of sentences in a set and the exact match for each submission. In cases where no majority agreement was reached, the answer provided by the highest-performing annotator was chosen.

\paragraph{Inter-Annotator Agreement.} We assessed the level of unanimous agreement on factors such as rewrite agreement (the number of sentences required to accurately rewrite a given sentence), exact match, and average Jaccard Similarity\footnote{Jaccard Similarity is defined as the size of the intersection of the sets divided by the size of the union of the sets.}. The initial annotation phase involved 64 native English speakers with a high approval rate (99\%) and significant experience on the AMT platform (over 5,000 completed HITs). Our analysis of the first 10\% of the data revealed rewrite agreement, exact match, and average Jaccard Similarity scores of 56\%, 67.5\%, and 94\%, respectively, with approximately 5\% of the data being unusable due to corrupted annotations. In order to continually improve the quality of our annotations, we narrowed our pool of annotators to the top five performers (based on activity and IAA performance) and provided personal feedback and bonuses based on the execution of each batch. As a result of these efforts, unanimous rewrite agreement, exact match, and average Jaccard Similarity all increased to 85\%, 82\%, and 97\%, respectively, and less than 1\% of the data required corrections.
\section{Conjunct Resolution Dataset} \label{sec:dataset_section}
Our Conjunct Resolution dataset consists of 10,206 verbal omission sentences, each paired with one of the conjunctions: "and", "or", and "but" (\cref{tab:dataset_conj_distribution} reports conjunction distribution) coupled with human annotations. By subtracting the number of verbs in the verbal omissions to those in the gold annotations, we find that 42\% of the verbs are omitted (see \cref{tab:dataset_verb_stats}). Furthermore, the majority (95.2\%) of sentences were found to be rewritable, with 82\% of the sentences being expressed in two sentences, 9.8\% being expressed in three sentences, 3.4\% expressed as four sentences or more, and only 4.8\% being classified as not rewritable.

\begin{table}[h]
\centering
\begin{tabular}{l|r|r|r}
Split & Explicit & Omitted & Total \\ 
\hline
Train  & 29,447 & 21,355 & 50,802\\
Validation & 3,630 & 2,631 & 6,261 \\
Test &  3,611 & 2,517 & 6,128\\
\hline\hline
Full &  36,688 & 26,502 & 63,190 \\
\end{tabular}
\caption{Count of explicit and omitted verbs in each split of the dataset, and the total count for each split. The full dataset contains a total of 10,206 instances}
\label{tab:dataset_verb_stats}
\end{table}
\begin{table}[h]
\centering
\begin{tabular}{l|r|r|r|r}
Split & and & or & but & Total\\ 
\hline
Train  & 6,508 &  798 & 860 & 8,166 \\
Validation & 805 & 108 & 108 & 1,021 \\
Test &  811 & 90 & 118 & 1,019 \\
\hline\hline
Full &  8,124 & 996 & 1,086 & 10,206  \\
\end{tabular}
\caption{Distribution of conjunctions (and, or, but) in each split of the dataset, and the total count for each split. The full dataset contains a total of 10,206 instances}
\label{tab:dataset_conj_distribution}
\end{table}

\paragraph{Non-rewritable Sentences.}
491 of the 10,206 (4.9\%) were marked as non-rewritable. Of these, 445 contain an explanation.\footnote{We did not collect explanations during the first few batches.} The reasoning behind deciding whether a sentence is rewritable or not seems to be non-trivial. For instance, consider the following two sentence:
\begin{enumerate}
\renewcommand*\labelenumi{(\theenumi)}
 \item I’d say Adam \textbf{will} \textbf{win} four majors and Justin \_\_ three, but I wouldn’t be surprised if it was the other way round.
 \item The pair are tied at the top after McIlroy \textbf{shot} 67 – his 25th score under par out of his last 27 rounds - and Horschel \_\_ a 69.
 \label{non_re_writable_example}
\end{enumerate}
Despite that both sentences are cases of gapping, only the second is rewritable. To recognize this, the reader (human or model) needs to be able to identify when two events are bound together by another piece of information. In the first sentence, the phrase "but I wouldn’t be surprised if it was the other way round" lacks context when appearing in the rewritten sentences. However, for the second sentence, there is no such issue, and is thus rewritable.\footnote{In linguistics and formal semantics, when a coordinated structure refers to the plurality of events as a whole, it is said to have a {\em collective} (as opposed to {\em distributive}) reading. The non-rewritable sentences in our set are those with collective readings. Their annotation and resolution is beyond the scope of this paper, and we reserve them for future work.} 
\section{Evaluation Metric} \label{sec:evaluation}
As no task is complete without an evaluation metric, we propose an automatic evaluation metric for the proposed conjunct resolution as rephrasing task.
For evaluation, we are interested in measuring three things: (1) how accurate the model is at resolving verbal omissions, (2) how often does the model omit other information after resolution, and (3) how often the model generates extra information. To address these three criteria, we propose to measure recall and precision over the \emph{predicate-argument relations} recovered by a dependency parser on the generated compared to the gold sentence set.

\paragraph{High Level Description.} The task revolves around making verbs and their arguments explicit. Our main object of interest is thus the ``verb nucleus'', an instance comprising of a verb and its arguments, as reflected in the dependency tree. We measure to what extent the nuclei extracted from the generated sentences overlap with the nuclei extracted from the gold sentences.
Neglecting to resolve an argument, or adding an extra argument to a given verb, will result in a mismatch between the gold and generated nucleus of that verb, hurting both recall and precision. Neglecting to spell out a verb completely will result in a missing nucleus (recall error), and over generating will result in spurious nuclei (precision error).

To calculate these metrics, we first produce dependency graphs for both the model's generated set of sentences and the gold annotation's set of sentences. From these graphs, we extract the verbs, and for each verb a subset of its dependents that we consider as arguments (based on a set of dependency labels). Each such set of verb+argument is a ``verb nucleus''. We treat the collection of verb nuclei over all sentences as a set (each element in the set is a collection of verb+arguments), and compute precision and recall over this set.

\paragraph{Details.} 
Denote the (automatically produced) syntactic dependency graphs of the $m$ gold sentences as $G = {\{g_1, g_2, \text{\dots}, g_m\}}$, and the dependency graphs of the $n$ generated sentences as $H = {\{h_1, h_2, \dots, h_n\}}$. We extract verb nuclei from these graphs, where each nucleus is a sub-graph containing a verb, its subject, object and lexicalized prepositional modifiers, as well as prepositional modifiers of the object and associated negations, if they exist.\footnote{The prepositional modifiers of the object are needed to handle pp-attachment errors of the parser. We refer the reader to \cref{app:evaluation_verb_nucleus} for a comprehensive account of the parser and included arguments.}
We represent a nucleus as a bag of $(w_1, \textit{dep}, w_2)$ triplets where $w_1$ and $w_2$ are words and $\textit{dep}$ is a dependency label, and consider two nuclei to be the same if their bags are the same. Denote by $N_{G}$ the bag of gold nuclei, by $N_{H}$ the bag of generated nuclei, and by $N_{I}$ the bag of identity nuclei, obtained by extracting nuclei from the input sentence. We obtain the subsets ${N'}_{H} = N_{H} \setminus N_{I}$ and ${N'}_{G} = N_{G} \setminus N_{I}$, which strictly contain nuclei with omitted verbs. Then $\textit{precision} = |{N'}_{H} \cap {N'}_{G}| \setminus |{N'}_{H}|$ and $\textit{recall} =|{N'}_{H} \cap {N'}_{G}| \setminus |{N'}_{G}|$. In cases where there is only one sentence in the gold set and the generated set, we skip the interaction with the identity nucleus, as to not punish the model for correctly not rewriting.

\paragraph{Calibration.} The input sentence contains some verbs and arguments that are repeated throughout the sentence. Thus, an approach that copies the full sentence two or more times will also yield some success under our metric, due to repeated verb nuclei, despite not being meaningful. To calibrate for this, we provide two additional numbers: (a) the precision and recall obtained by a model that spits the original sentence as output, unmodified; and (b) the precision and recall of a model that has access to the correct number $k$ of sentences in a gold annotation set, and which uses this information by spitting out $k$ copies of the input sentence. Model performance should always be judged in comparison to these baselines.

\begin{figure}[H]
    \centering
    \includegraphics[width=\columnwidth]{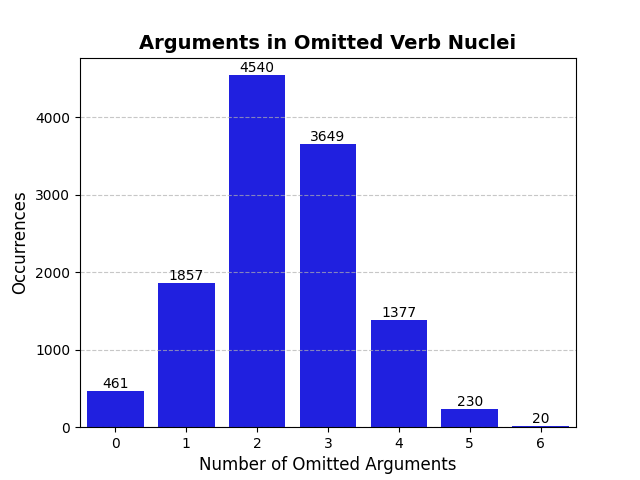}
    \caption{A distribution over the number of omitted arguments in a verb nucleus}
    \label{fig:omitted_nuclei}
\end{figure}
\section{Experiments}
We evaluate neural models on the task, both in supervised fine-tuning and in in-context learning setups. For the supervised fine-tuning case, we measure both task performance as well as the dependence of performance on dataset size. As a concluding experiment, we take the best-performing model and manually evaluate it against the task definition.

\paragraph{Dataset Split and Preprocessing.} We shuffle and then split our dataset to train (80\%), validation (10\%), and test (10\%) sets. 
Each input instance contains a sentence and a marked conjunction.
Each output instance is a sequence of sentences. 
For feeding the text to the models, we mark the conjunction using a special token\footnote{In T5, sentinel tokens were employed, and in GPT3, the "<SPLIT>" marker was utilized.} and separate the output sentences using special tokens.\footnote{In T5, sentinel tokens were employed again, and in GPT3, each sentence was written in a separate line.}

\paragraph{Supervised Fine-tuning.} 
We fine-tune ten T5-large \citep{t5} models, using increasingly larger subsets of the training data, from 10\% to 100\% in increments of ten.
All models were trained with the same hyperparameters and fine-tuned for five epochs, with the best performing model on the validation set being saved and subsequently evaluated on the test-set (for the detailed training configuration, see \labelcref{app:models}).

\paragraph{In-context Learning / Prompting.} 
We evaluate the state-of-the-art GPT \texttt{text-davinci-3} model from OpenAI, in an in-context learning (prompting) fashion \citep{gpt3}.

To obtain in-context examples, we randomly sampled for each test instance three re-writable and one non-rewritable sentence, sharing the conjunction with the test instance. Further details of the prompt and parameters are available in the appendix \labelcref{app:models}.

\paragraph{Manual Evaluation.} Our evaluation metric cannot fully evaluate the semantic correctness of results. To overcome this, we perform a manual evaluation in which a human annotator is requested to examine the system's inputs (sentences containing omissions) together with outputs of the best performing-model, and assess whether the generated set of sentences has the same meaning as the input sentence, or a different one.

\subsection{Results and Discussion}
The results of all models are summarized in \cref{tab:baseline_results}.
Results improve rapidly, but then saturate on around 82\% F1 already with 40\% ($\sim3,200$) training samples, reaching a peak of 82.4\% F1, indicating that the key to the task may not be ``more data''.

In our experiments, the $T5_{40\%}$ and $T5_{80\%}$ models demonstrated the best performance. However, the $T5_{80\%}$ model had a more balanced performance across the different metrics, making it the preferred model for further analysis and interpretation of results.
When examining performance on specific conjunctions, the best performing model, $T5_{80\%}$, scored 83.7\% on ``and'' sentences, 76.3\% on ``or'' sentences, and on ``but'' sentences, it scored 77.7\% F1. $GPT3$ performed similarly, but to a lesser extent, scoring 73.5\%, 70.4\%, and 65.4\% F1 on average, in the aforementioned order.

In terms of quantity, at a minimum, models learned that the resolution is centered around the verb. $T5_{80\%}$ generates 5,514 out of the 6,128 (89.9\%) verbs in the gold annotations, while only over-generating 349 verbs, a mere 5.7\% of the total verbs in the test set. Similarly, $GPT3$ generates 4,680 out of the 6,128 (75\%) and over-generates 273 verbs (4.4\%).

Finally, the manual evaluation evaluating the semantic correctness of the $T5_{80\%}$ system's outputs with respect to the input, reveals that in 87.9\% of the cases the meaning is preserved, and in 12.1\% it is not. Although measuring different aspects of the answer, these numbers are similar to the automatic F1 results, establishing some additional trust in it as an automated metric.

\begin{table}[t]
\centering
\begin{tabular}{l|ccc}
\textbf{Model} & \textbf{Recall} & \textbf{Precision} & \textbf{F1}\\
\hline
Calibration$_{1}$ & 5.1 & 5.1 & 5.1\\
Calibration$_{k}$ & 49.8 & 41.8 & 45.5\\
\hline
$T5_{10\%}$ & 75.9 & 43.2 & 55.1\\
$T5_{20\%}$ & 77.2 & 78.2 & 77.7\\
$T5_{30\%}$ & 79.5 & 79.6 & 79.5\\
$T5_{40\%}$ & \textbf{82.4} & 81.8 & 82.1\\
$T5_{50\%}$ & 81.6 & 82.1 & 81.8\\
$T5_{60\%}$ & 81.2 & 81.5 & 81.3\\
$T5_{70\%}$ & 81 & 82.2 & 81.6\\
$T5_{80\%}$ & 82 & \textbf{82.7} & \textbf{82.3}\\
$T5_{90\%}$ & 81.2 & 82.1 & 81.6\\
$T5_{100\%}$ & 81.8 & \textbf{82.7} & 82.2\\
\hline
$GPT3$ & 68.7 & 76.2 & 72.3\\
\hline
\end{tabular}
\caption{Comparison of model performance in terms of recall, precision, and F1 score on the test set. The T5 models trained on 40\% of the training set or higher generally perform similarly, with $T5_{80\%}$ achieving the highest overall F1 score of 82.3. The GPT3 model has a lower overall performance, with an F1 score of 72.3. Calibration$_1$ and Calibration$_k$ are the results of models that output back the input sentence, or $k$ instances of the input sentence, where $k$ is the number of sub-sentences in the rewritten text.}
\label{tab:baseline_results}
\end{table}

\subsection{Error Examples}
Below are some examples of errors by the best-performing T5 model.

\noindent \textbf{Attributing the correct arguments to a different verb.} Here, the gold annotation indicates the verb ``playing'' as the omission, while the model wrongly chose ``gain'' (crossed items are missing in output):

\vspace{5pt}

 \noindent \textit{Players \emph{gain} points from \emph{playing} in the four grand slams, ATP World Tour events including eight Masters Series tournaments, and World Group singles matches in the Davis Cup teams competition.}
\begin{enumerate}
        \renewcommand*\labelenumi{(\alph{enumi})}
         \item[] (a) \textit{Players gain points from  playing in the four grand slams.}
         \\  (b) \textit{Players gain points from \st{playing in} ATP World Tour events including eight Masters Series tournaments.}
         \\ (c) \textit{Players gain points from \st{playing in} World Group singles matches in the Davis Cup teams competition.}
\end{enumerate}        

\noindent \textbf{Incorrectly attaching two arguments to the same verb.} Here the model should have produced a different sentence for ``over 300,000 injured'':

\vspace{5pt}

\noindent\textit{The January 12 quake left over 220,000 dead, over 300,000 injured, and over a million homeless.}

       \begin{enumerate}
        \renewcommand*\labelenumi{(\alph{enumi})}
         \item[] (a) \textit{The January 12 quake left over 220,000 dead, \textbf{over 300,000 injured}.}
         \\  (b) \textit{The January 12 quake left over a million homeless.}
        \end{enumerate}   

\noindent\textbf{Failure to resolve verb-phrase ellipsis.}

\vspace{5pt}

\noindent\textit{Germany knows how to do it, and we don’t.}
\begin{enumerate}
    \renewcommand*\labelenumi{(\alph{enumi})}
         \item[] (a) \textit{Germany knows how to do it.}
         \\ (b) \textit{We don't \st{know how to do it}.}
\end{enumerate}   

\noindent\textbf{Considering a different coordination.} Here, the model split around `;' rather than around `and', missing information in the process: 

\vspace{5pt} 

\noindent\textit{Johnston's one shot went through her front door and over the officers' heads; they responded with 39 shots, hitting Johnston five times.}
   \begin{enumerate}
    \renewcommand*\labelenumi{(\alph{enumi})}
     \item[] (a) \textit{Johnston's one shot went through her front door \_\_.}
     \\ (b) \textit{the officers responded with 39 shots, hitting Johnston five times.}
    \end{enumerate}

\noindent\textbf{Missing conjuncts from the resolution.} Here, the crossed item did not appear in the model's output, but should have: 

\vspace{5pt} 

\noindent\textit{At the 2001 Census, 92.4\% of the city's populace was White—including one percent White Irish—3.8\% were South Asian, 1.0\% Black, 1.3\% Chinese or other ethnic groups, and 1.5\% were of Mixed Race.}
   \begin{enumerate}
    \renewcommand*\labelenumi{(\alph{enumi})}
     \item[] (a) \textit{At the 2001 Census, 92.4 \% of the city's populace was White—including one percent White Irish.}
    \\ (b) \textit{\st{At the 2001 Census, 3.8 \% of the city's populace was South Asian.}}
     \\ (c) \textit{At the 2001 Census, 1.0 \% of the city's populace was Black.}
     \\ (d) \textit{At the 2001 Census, 1.3 \% of the city's populace was Chinese or other ethnic groups.}
     \\ (e) \textit{At the 2001 Census, 1.5 \% of the city's populace was of Mixed Race.}
    \end{enumerate}   
    
\noindent\textbf{Resolving creates a factual inaccuracy.} Here, the model splits a sentence to create factually incorrect sentences: 

\vspace{5pt} 

\noindent\textit{Heist had been sentenced to three years of probation for the identity theft and for giving false information to a law enforcement officer.}
   \begin{enumerate}
    \renewcommand*\labelenumi{(\alph{enumi})}
     \item[] (a) \textit{Heist had been sentenced to three years of probation for the identity theft.}
     \\ (b) \textit{Heist had been sentenced to three years of probation for giving false information to a law enforcement officer.}
    \end{enumerate}   

This error example illustrates an inherent limitation of our approach, as the correct semantics cannot be represented as a set of sentences. The correct behavior under our representation would have been to not split this sentence at all.
\section{Conclusions}
We present a novel approach for studying verbal omissions in coordination structures. Previous research in this area has been fragmented, focusing on individual phenomena. In contrast, we propose a unified approach which considers all  conjunction related verbal omissions under the same framework, by introducing a text-to-text conjunct-resolution task, to resolve omitted verbs and their arguments.  
We compiled and curated a large dataset of conjunction related verbal omissions, consisting of over 10,000 sentences and human annotations, which serves as a valuable resource for further research in this area.
Our results using state-of-the-art models as neural baselines demonstrate that this task is challenging and merits further work.

\section*{Limitations}
One unsatisfying aspect of proposed task is that it accounts for {\em distributive} coordination structures, but is not able to handle sentences with {\em collective} reading where the main predicate applies to the plurality of conjuncts as a whole. In our data collection these account for about 4.9\% of the verbal omission cases, and such sentences are left ``non-rewritable''. In future work, we would like a solution that allows to resolve also such sentences in a consistent yet easy-to-annotate manner.

Additionally, in the GPT prompting experiment we experimented with a few different prompts, but did not do exhaustive prompt engineering, and it is possible that with more aggressive prompt engineering GPT can perform better on the task than our results indicate. Similarly for the fine-tuning experiments with T5-large, in which we did some hyperparameter tuning, but not aggressively so.

\section*{Ethics Statement}
\paragraph{Worker Qualification and Compensation for Annotation.}
To collect annotations on our dataset, we used Amazon Mechanical Turk (AMT). All workers had the following qualifications: (1) over 5,000 completed HITs; (2) 99\% approval rate or higher; (3) Native English speakers from England, New Zealand, Canada, Australia, or United States. Workers were paid \$0.75 per HIT, and on average completed a batch within four hours of work. In addition, \$10 was given upon completing a batch (73 HITs), raising the hourly pay to ~\$16.2.

\paragraph{Data Collection and Usage Policy for Annotation.}
Workers were informed that their annotations would be collected for research purposes and would be used to train and evaluate language-related models, and that the annotations would eventually be made publicly available. Additionally, our task and the annotations collected were of objective nature and did not contain any personal information. Furthermore, all data sources used in the study were publicly available.

\section{Acknowledgements} 
This project received funding from the Europoean Research Council (ERC) under the Europoean Union's Horizon 2020 research and innovation programme, grant agreement No.\ 802774 (iEXTRACT) and grant agreement No.\ 677352 (NLPRO). The  third author is also  funded by a grant from the Israeli Ministry of Science and Technology (MOST), grant number 3-17992.

\bibliography{anthology,custom}
\bibliographystyle{acl_natbib}
\clearpage
\newpage

\appendix

\section{Appendix}
\label{sec:appendix}

\subsection{Parser}
\label{app:parser_used}
Throughout this project, we use the spacy parser with the out-of-the-box \texttt{en\_core\_web\_trf} model.

\subsection{Models} \label{app:models}
\paragraph{T5 Configuration.}
To ensure reproducibility of our results, we provide the specific configuration of T5-large model used in our study. The model had 770M parameters and was trained with a batch size of 8. The optimizer used was AdamW with a adam\_eps value of 1e-8. The maximum input and output length were set to 256 and weight decay was set to 0. The learning rate was set according to the recommendations of the Hugging Face library, with a value of 3e-4. These configurations were used consistently across all variations of the model in our study.

\paragraph{GPT-3 Configuration.}
In our approach, we use OpenAI's \texttt{text-davinci-3} model, a large language model based on the GPT-3 architecture. The temperature is set to 0 and top\_p remains 1, resulting in conservative and less-deviant text. The maximum number of tokens generated is set to 256. These parameters have been fine-tuned to control the generated text.

\paragraph{GPT-3 Prompt Format.}Here we show a concrete illustration of the instructions provided to GPT-3.\\

\textit{Q: 58.1\% of the population described themselves in the 2011 census return as being at least nominally Christian \textbf{<SPLIT> and </SPLIT>} 0.7\% as Muslim with all other religions represented by less than 0.5\% each.}\\
\textit{A:}\\
\textit{Cannot re-write this sentence.}\\

\noindent\textit{Q: Federal education assistance offered affordable loans to Americans who wanted to attend college \textbf{ <SPLIT> and </SPLIT>} money for local schools to ensure that all children received an adequate education.}\\
\textit{A:}\\
\textit{Federal education assistance offered affordable loans to Americans who wanted to attend college.}\\
\textit{Federal education assistance offered money for local schools to ensure that all children received an adequate education.}\\

\noindent\textit{Q: He was subsequently asked to repeat the program at the American Asylum for Deaf - mutes in Hartford, Connecticut, \textbf{<SPLIT> and </SPLIT>} the Clarke School for the Deaf in Northampton, Massachusetts.}\\
\textit{A:}\\
\textit{He was subsequently asked to repeat the program at the American Asylum for Deaf - mutes in Hartford, Connecticut.}\\
\textit{He was subsequently asked to repeat the program at the Clarke School for the Deaf in Northampton, Massachusetts.}\\

\noindent\textit{Q: The plan, a grid with two main axes meeting at a central square \textbf{<SPLIT> and </SPLIT>} an additional square in each corner, was based on Thomas Holme's 1682 plan for Philadelphia.}\\
\textit{A:}\\
\textit{The plan, a grid with two main axes meeting at a central square, was based on Thomas Holme's 1682 plan for Philadelphia.}\\
\textit{The plan, a grid with an additional square in each corner, was based on Thomas Holme's 1682 plan for Philadelphia.}\\

\textit{Q: Example alternative schools include Montessori schools, Waldorf schools, Friends schools, Sands School, Summerhill School, The Peepal Grove School, Sudbury Valley School, Krishnamurti schools, \textbf{<SPLIT> and </SPLIT>} open classroom schools.}

\subsection{Crowdsourcing Task} \label{app:crowdsourcing_task}
Here we show the instructions in our crowdsourcing annotation task.

\begin{figure}[h!]
    \centering
    \includegraphics[width=\columnwidth]{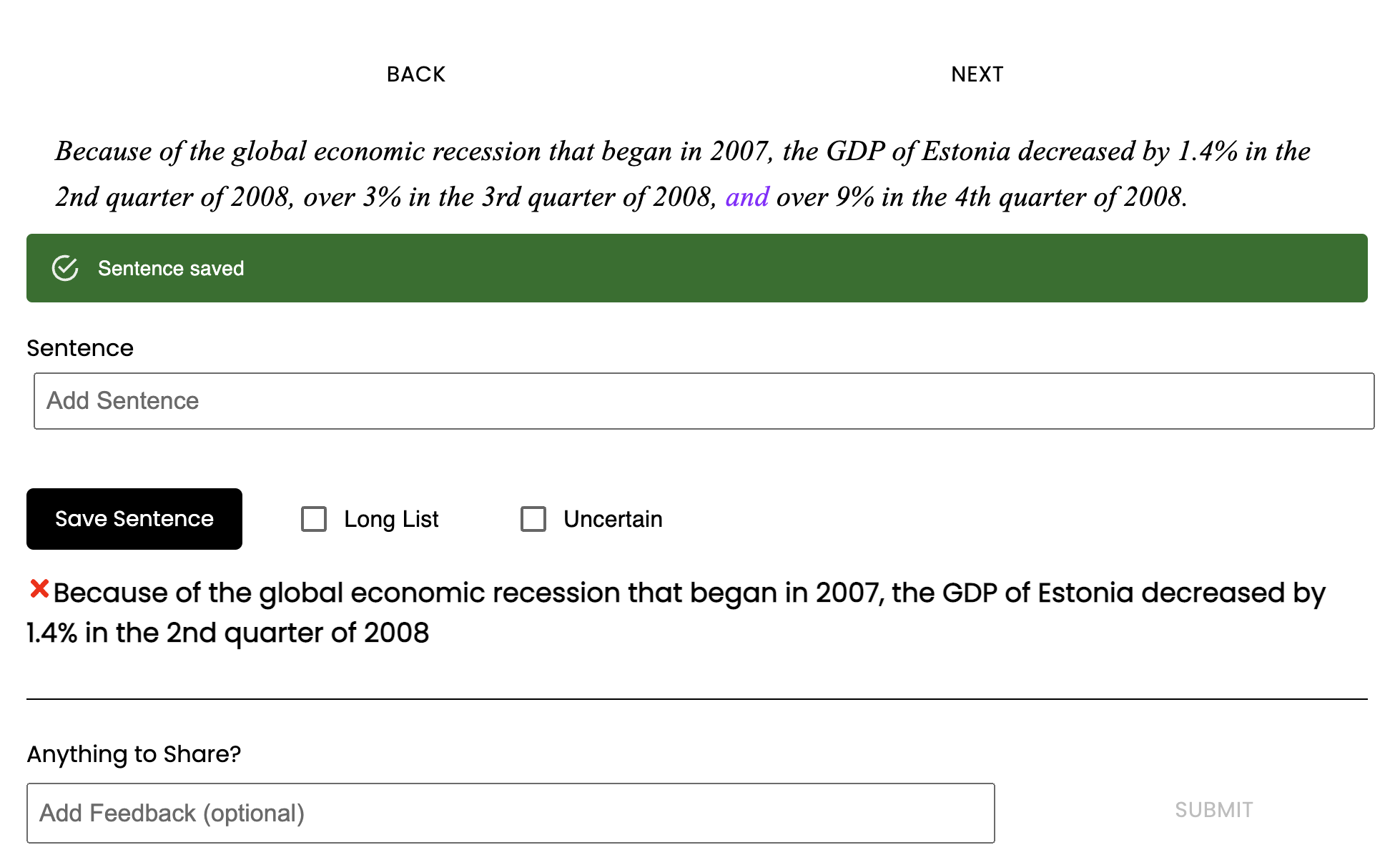}
    \caption{Crowdsourcing task UI}
    \label{fig:hit_submit}
\end{figure}

\clearpage

\subsection{Crowdsourcing Task Instructions}
\scalebox{0.95}{
    \includegraphics[width=\textwidth]{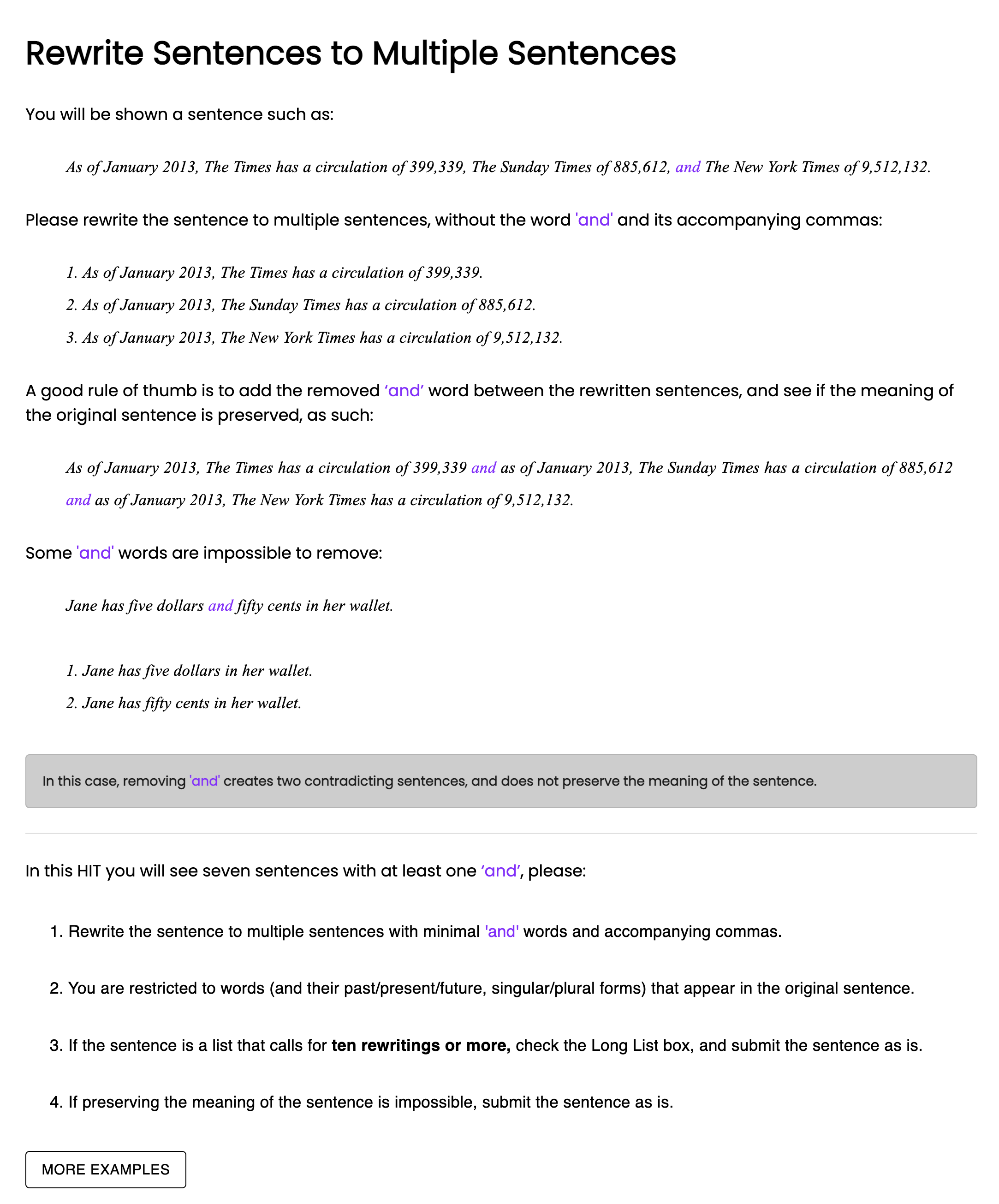}
    \label{fig:hit_instructions}
}
\clearpage

\subsection{Dependencies in Verb Nucleus.} 
\label{app:evaluation_verb_nucleus}
As detailed in \cref{sec:evaluation}, a verb nucleus contains a verb and its arguments. While to identify the nucleus root (the verb), we 
look if their part-of-speech tag is one of (``VB'', ``VBD'', ``VBG'', ``VBN'', ``VBP'', ``VBZ''), the rest of the nucleus is defined over the dependency graph:
\begin{itemize}
\item subjects -- (``nsubj'', ``nsubjpass'', ``expl'').

\item object -- (``dobj'', ``obj'', ``pobj'', ``iobj'', ``attr'', ``oprd'').

\item prepositions and their prepositional modifiers -- (``prep'', ``agent'') and (``pobj'', ``pcomp'').

\item negations -- ``neg''.
\end{itemize}

Negations are included to account for cases where the model correctly predicts most verb arguments, but fails to account for negation, thus, breaking the original meaning of the sentence. For instance, ``The governor urged the public not to panic and to follow his reports closely'' is resolved to:

\begin{itemize}
\item \textit{The governor urged the public not to panic}
\item \textit{The governor urged the public \st{not} to follow his reports closely}
\end{itemize}

\subsection{Additional Results}
To put the performance of the models in context, we provide results over each conjunct. Moreover, we include exact matching over the sets of sentences, here, punctuation is removed and the sets are assumed to be aligned. See \cref{app:conj_based_results}.

\begin{table}[H]
\centering
\scalebox{0.8}{

\begin{tabular}{c|c|c|c|c|c}
\textbf{Model} & \textbf{F1$_{\text{and}}$} & \textbf{F1$_{\text{but}}$} & \textbf{F1$_{\text{or}}$} & \textbf{F1} & \textbf{Exact Match} \\
\hline
Calibration$_{1}$ & 4.7 & 0.0 & 15.6 & 5.1 & 2.4 \\
Calibration$_{k}$ & 45.2 & 30.9 & 64.9 & 45.5 & 2.4 \\
\hline
$T5_{10\%}$  & 59.7 & 28.9 & 45.2 & 55.1 & 37.3 \\
$T5_{20\%}$  & 81.2 & 57.7 & 72.2 & 77.7 & 66.6 \\
$T5_{30\%}$  & 82.9 & 60.4 & 74.2 & 79.5 & 69.1 \\
$T5_{40\%}$  & \textbf{84.3} & 70.9 & 76.3 & 82.1 & 72.4 \\
$T5_{50\%}$  & 83.5 & 75.7 & 74.2 & 81.8 & 72.6 \\
$T5_{60\%}$  & 82.6 & 76.2 & 76.9 & 81.3 & 72.7 \\
$T5_{70\%}$  & 82.8 & 76.4 & 76.7 & 81.6 & 73.8 \\
$T5_{80\%}$  & 83.7 & 77.7 & 76.3 & \textbf{82.3} & 73.9 \\
$T5_{90\%}$  & 82.9 & 77.5 & 75.3 & 81.6 & 72.6 \\
$T5_{100\%}$ & 82.6 & \textbf{79.3} & \textbf{82.4} & 82.2 & \textbf{74.9} \\
\hline
$GPT3$ & 73.5 & 65.4 & 70.4 & 72.3 & 53.1 \\
\hline
\end{tabular}

}
\caption{F1 performance of all models based on their conjunction. For context, an exact match over the sets of sentences is provided.}
\label{app:conj_based_results}
\end{table}


\subsection{Additional Examples of Patterns Indicating Suspicious Sentences}
In collecting sentences, we broadly look for three types of "suspicious" structures:
\begin{itemize}
\item  \textbf{\emph{part-of-speech mismatch}} -- two words with a different part-of-speech are linked by a \textit{conj}.
\item \textbf{\emph{dependency relation mismatch}} -- an inconsistency between a word's part-of-speech tag and its relations to its dependents.
\item \textbf{\emph{subtree mismatch}} -- two words with the same part-of-speech, but different subtrees.
\end{itemize}

While most patterns involve more than one suspicious structure, we group the examples to match the above list.

The full sentences demonstrating these patterns can be seen below. We denote common-nouns, proper-nouns, adjectives, and numerical values with ``NON-VERB'', and a verb or an auxiliary verb, with ``VERB''. When a pattern checks for one of few possible relations between two words, we use ``/'' to separate them (e.g., \textit{advcl/xcomp} indicates the pattern accepts an \textit{advcl} or an \textit{xcomp} relation). The relation \textit{any} indicates the pattern accepts any type of relation between two words, and \textit{obj} indicates the pattern accepts any type of object.

\subsubsection{Part-of-speech Mismatch}
\vspace{1cm}
\tikzset{every node/.style={align=center}}
\begin{dependency}
\begin{deptext}[column sep=0.5em, row sep =.5em]
\dots \& * \& \dots \& * \&  \dots \\
       \& VERB  \&      \&     NON-VERB \& \\
\end{deptext}
\depedge[edge height=5ex]{2}{4}{conj}
\end{dependency}\\
\noindent\textbf{Example sentence:} Koreans made up 1.2\% of the city's population, and Japanese 0.3\%.\\

\tikzset{every node/.style={align=center}}
\begin{dependency}
\begin{deptext}[column sep=0.5em, row sep =.5em]
\dots \& * \& \dots \& * \& \dots \\
        \& VERB       \&    \&  NON-VERB     \\
\end{deptext}
\depedge[edge height=5ex]{2}{4}{conj}
\end{dependency}\\
\noindent\textbf{Example sentence:} Ten chapters are devoted to body issues and how to
cover them.\\

\clearpage
\tikzset{every node/.style={align=center}}
\begin{dependency}
\begin{deptext}[column sep=0.5em, row sep =.5em]
\dots \& * \& \dots \& * \&\dots \\
       \& ADV \&        \&     ADP \&          \\
\end{deptext}
\depedge[edge height=5ex]{2}{4}{conj}
\end{dependency}\\
\noindent\textbf{Example sentence:} Desormeaux has won the Preakness twice: once
aboard Real Quiet in 1998 and again 10 years later
on Big Brown.\\

\tikzset{every node/.style={align=center}}
\begin{dependency}
\begin{deptext}[column sep=0.5em, row sep =.5em]
\dots \& *     \& \dots \& * \& \dots \\
 \& ADP/PART   \&            \& NON-VERB  \& \\ 
\end{deptext}
\depedge[edge height=5ex]{2}{4}{conj}
\end{dependency}\\
\noindent\textbf{Example sentence:} Tell us in the comments below or @CNNFilms on
Twitter.\\

\tikzset{every node/.style={align=center}}
\begin{dependency}
\begin{deptext}[column sep=0.5em, row sep =.5em]
\dots \& * \& \dots \& * \& \dots \\
      \&  ADJ \&     \& ADP    \&    \\
\end{deptext}
\depedge[edge height=5ex]{2}{4}{conj}
\end{dependency}\\
\noindent\textbf{Example sentence:} Southwest said all customers were safe and at the
terminal.\\

\subsubsection{Dependency Relation Mismatch}
\vspace{1cm}

\tikzset{every node/.style={align=center}}
\begin{dependency}
\begin{deptext}[column sep=0.5em, row sep =.5em]
\dots \& * \& * \& \dots \\
       \& NON-VERB   \& NON-VERB     \&       \\
\end{deptext}
\depedge[edge height=1.5ex]{3}{2}{nsubj}
\end{dependency}\\
\noindent\textbf{Example sentence:} To idealists, spirit or mind or the objects of mind are
primary, and matter secondary.\\


\subsubsection{Subtree Mismatch}
\vspace{1 cm }

\tikzset{every node/.style={align=center}}
\scalebox{0.7}{
\begin{dependency}
\begin{deptext}[column sep=0.5em, row sep =.5em]
\dots \& * \& \dots \& * \& \dots \& * \& \dots \& * \& \dots \\
        \& VERB    \&       \&  CCONJ \&       \&  ADP  \& \& ADP \& \\
\end{deptext}
\depedge[edge height=1.5ex]{2}{8}{prep}
\end{dependency}
}
\noindent\textbf{Example sentence:} John was born to Henry II of England and Eleanor of
Aquitaine on 24 December 1166.\\

\tikzset{every node/.style={align=center}}
\begin{dependency}
\begin{deptext}[column sep=0.5em, row sep =.5em]
* \& \dots \& * \& \dots \& * \\
 ADP \&           \& ADV     \&      \& VERB\\
\end{deptext}
\depedge[edge height=1.5ex]{1}{3}{advmod}
\depedge[edge height=3.ex]{1}{5}{prep}
\end{dependency}\\
\noindent\textbf{Example sentence:} From the 1880s onward neighbourhoods such as Oud-
wijk, Wittevrouwen, Vogelenbuurt to the East, and
Lombok to the West were developed.\\

\tikzset{every node/.style={align=center}}

\begin{dependency}
\begin{deptext}[column sep=0.5em, row sep =.5em]
\dots \& * \& *  \& \dots \& * \& \dots \& * \& \dots \\
       \&   \&    \&      \&    \&   \& VERB \&  \\
\end{deptext}
\depedge[edge height=2.5ex]{2}{3}{xcomp}
\depedge[edge height=5ex]{2}{5}{conj}
\depedge[edge height=5ex]{5}{7}{conj}
\end{dependency}\\
\noindent\textbf{Example sentence:} 19 soldiers, policemen reported wounded, and some
attackers killed, wounded or captured.\\

\tikzset{every node/.style={align=center}}
\scalebox{0.8}{
\begin{dependency}
\begin{deptext}[column sep=0.5em, row sep =.5em]
\dots \& * \& * \& \dots  \& * \& \dots \& * \& \dots \\
      \& VERB    \&              \&     \&           \&      \&  NON-VERB           \& \\
\end{deptext}
\depedge[edge height=1.5ex]{2}{3}{obj}
\depedge[edge height=1.5ex]{7}{5}{prep/agent}
\depedge[edge height=5ex]{3}{7}{conj}
\end{dependency}}\\
\noindent\textbf{Example sentence:} You send out these sound waves, and when they
bounce off of objects, the reflection of the waves
tells you – or in this case, the animal – where the
objects are.\\

\tikzset{every node/.style={align=center}}
\scalebox{0.7}{
\begin{dependency}
\begin{deptext}[column sep=0.5em, row sep =.5em]
\dots \& * \& * \& * \& * \& \dots \& * \& * \& \dots \\
      \&         \&         \&     \&      \&      \& PART \& VERB \& \\
\end{deptext}
\depedge[edge height=1.5ex]{2}{3}{any}
\depedge[edge height=1.5ex]{3}{4}{any}
\depedge[edge height=1.5ex]{4}{5}{any}
\depedge[edge height=5ex]{5}{8}{conj}
\depedge[edge height=1.5ex]{8}{7}{aux}
\end{dependency}}\\
\noindent\textbf{Example sentence:} Some runners started raising money for charity or to
help with relief efforts.\\

\tikzset{every node/.style={align=center}}
\scalebox{0.8}{
\begin{dependency}
\begin{deptext}[column sep=0.5em, row sep =.5em]
\dots \& * \& * \& * \& \dots \& * \& * \& \dots \\
      \& VERB          \&    \&        \&      \& NON-VERB    \&   \& \\
\end{deptext}
\depedge[edge height=3.5ex]{2}{3}{agent/prep}
\depedge[edge height=1.5ex]{3}{4}{pobj/pcomp}
\depedge[edge height=5ex]{4}{6}{conj}
\depedge[edge height=1.5ex]{6}{7}{prep}
\end{dependency}
}\\
\noindent\textbf{Example sentence:} Every day, someone new is introduced to the hardships of wartime military service or the horrors of
combat.\\

\tikzset{every node/.style={align=center}}
\scalebox{0.8}{
\begin{dependency}
\begin{deptext}[column sep=0.5em, row sep =.5em]
* \& \dots \&  * \& \dots \& * \& * \\
NON-VERB \&       \&    NON-VERB      \&       \&    \&  \\
\end{deptext}
\depedge[edge height=1ex]{6}{5}{subj}
\depedge[edge height=3ex]{1}{6}{ccomp}
\depedge[edge height=5ex]{1}{3}{conj}
\end{dependency}}\\
\noindent\textbf{Example sentence:} Progress in the Business District but lingering blight
in poorer neighborhoods, he says.\\

\tikzset{every node/.style={align=center}}
\scalebox{0.8}{
\begin{dependency}
\begin{deptext}[column sep=0.5em, row sep =.5em]
\dots \& * \& * \& * \& *  \& * \& * \& *\\
       \& VERB \&         \&                \&     \&  NON-VERB           \&     \&       \\
\end{deptext}
\depedge[edge height=1.5ex]{2}{3}{obj}
\depedge[edge height=5ex]{3}{6}{conj}
\depedge[edge height=1.5ex]{3}{4}{prep}
\depedge[edge height=1.5ex]{4}{5}{pobj}
\depedge[edge height=1.5ex]{6}{7}{prep}
\depedge[edge height=1.5ex]{7}{8}{pobj}
\end{dependency}
}\\
\noindent\textbf{Example sentence:} In 1995, material costs were 30 cents for the jewel
case and 10 to 15 cents for the CD.\\

\tikzset{every node/.style={align=center}}
\scalebox{0.8}{
\begin{dependency}
\begin{deptext}[column sep=0.5em, row sep =.5em]
\dots \& * \& \dots  \& * \& * \& * \& \dots \\
       \& VERB          \&    \&                    \&       \&         \\
\end{deptext}
\depedge[edge height=5ex]{2}{4}{conj}
\depedge[edge height=1.5ex]{4}{5}{prep}
\depedge[edge height=1.5ex]{5}{6}{pobj}
\end{dependency}}\\
\noindent\textbf{Example sentence:} Neesham would make 85 from 80 and Kane
Williamson a more considered 54 from 98 as Sri
Lanka toiled.\\

\tikzset{every node/.style={align=center}}
\scalebox{0.85}{
\begin{dependency}
\begin{deptext}[column sep=0.5em, row sep =.5em]
\dots \& * \& * \& \dots \&  * \& \dots \\
      \&   VERB        \&     \&     \&  ADP \&     \\
\end{deptext}
\depedge[edge height=1.5ex]{2}{3}{prep}
\depedge[edge height=5ex]{3}{5}{conj}
\end{dependency}}\\
\noindent\textbf{Example sentence:} It is also used in woodcut printmaking, and for engraving.\\

\tikzset{every node/.style={align=center}}
\scalebox{0.8}{
\begin{dependency}
\begin{deptext}[column sep=0.5em, row sep =.5em]
\dots \& * \& * \& * \& \dots \& \dots \& * \& \dots \\
 \&   VERB   \&                 \&   \&     \&       \& ADP \&     \\
\end{deptext}
\depedge[edge height=1.5ex]{2}{3}{obj}
\depedge[edge height=1.5ex]{3}{4}{prep}
\depedge[edge height=5ex]{4}{7}{conj}
\end{dependency}
}\\
\noindent\textbf{Example sentence:} This is The Joker’s war on Batman and even more so,
on his family.\\

\tikzset{every node/.style={align=center}}
\scalebox{0.75}{
\begin{dependency}
\begin{deptext}[column sep=0.5em, row sep =.5em]
\dots \& * \& * \& * \& * \& * \& \dots \&  * \& \dots \\
      \&   VERB     \&              \&       \&   \&       \&     \&  ADP \& \\
\end{deptext}
\depedge[edge height=1.5ex]{2}{3}{obj}
\depedge[edge height=1.5ex]{3}{4}{any}
\depedge[edge height=1.5ex]{4}{5}{any}
\depedge[edge height=1.5ex]{5}{6}{prep}
\depedge[edge height=5ex]{6}{8}{conj}
\end{dependency}
}\\
\noindent\textbf{Example sentence:} They’ve been major players in the uprisings in Yemen
and in Syria.\\

\tikzset{every node/.style={align=center}}
\scalebox{0.65}{
\begin{dependency}
\begin{deptext}[column sep=0.5em, row sep =.5em]
* \& * \& \dots \&  * \& \dots \& * \& \dots \\
     NON-VERB   \&        \&     \&  ADP  \&       \&     VERB        \&       \\
\end{deptext}
\depedge[edge height=1.5ex]{1}{2}{prep}
\depedge[edge height=6ex]{2}{4}{conj}
\depedge[edge height=3ex]{6}{1}{subj}
\end{dependency}
}\\
\noindent\textbf{Example sentence:} Government control of the economy and of expression is much reduced, he says.\\

\tikzset{every node/.style={align=center}}
\scalebox{0.7}{
\begin{dependency}
\begin{deptext}[column sep=0.5em, row sep =.5em]
\dots \& * \& * \& \dots \& *  \& \dots \& * \& \dots \\
      \&   VERB        \&        \&           \&     \&       \& ADP \&     \\
\end{deptext}
\depedge[edge height=1.5ex]{2}{3}{prep}
\depedge[edge height=1.5ex]{3}{5}{pobj}
\depedge[edge height=5ex]{5}{7}{conj}
\end{dependency}}\\
\noindent\textbf{Example sentence:} They concentrated in trade, services, and especially
in money lending.\\

\tikzset{every node/.style={align=center}}
\scalebox{0.65}{
\begin{dependency}
\begin{deptext}[column sep=0.5em, row sep =.5em]
\dots \& * \& * \& \dots  \& * \& * \& \dots \\
        \& VERB \& NON-VERB    \&    \& NON-VERB        \& NON-VERB  \&    \\
\end{deptext}
\depedge[edge height=1.5ex]{2}{3}{obj}
\depedge[edge height=5ex]{3}{6}{conj}
\depedge[edge height=1.5ex]{6}{5}{compound}
\end{dependency}
}
\noindent\textbf{Example sentence:} The meteor show is entertainment for most, but a
research chance for NASA.\\
\end{document}